\documentclass[letterpaper, 10 pt, conference]{ieeeconf}  % Comment this line out if you need a4paper

\usepackage{graphicx}
\usepackage{subcaption}
\usepackage{amsmath}
\usepackage{multirow}

\IEEEoverridecommandlockouts                              % This command is only needed if 
\title{\LARGE \bf
Dynamic Recalibration in LiDAR SLAM: Integrating AI and Geometric Methods with Real-Time Feedback Using INAF Fusion*
}

\author{Zahra Arjmandi$^{1}$ and Gunho Sohn$^{1}$% <-this % stops a space
\thanks{*This work was not supported by any organization}% <-this % stops a space
\thanks{$^{1}$Zahra Arjmandi and Gunho Sohn are with the Department of Earth and Space Science and Engineering, Lassonde School of Engineering, York University, 4700 Keele Street, Toronto, Ontario M3J 1P3, Canada.
        {\tt\small zahraarj@yorku.ca, gsohn@yorku.ca}}%
}

\begin{document}

\maketitle
\thispagestyle{empty}
\pagestyle{empty}

%%%%%%%%%%%%%%%%%%%%%%%%%%%%%%%%%%%%%%%%%%%%%%%%%%%%%%%%%%%%%%%%%%%%%%%%%%%%%%%%
\begin{abstract}

This paper presents a novel fusion technique for LiDAR Simultaneous Localization and Mapping (SLAM), aimed at improving localization and 3D mapping using LiDAR sensor. Our approach centers on the Inferred Attention Fusion (INAF) module, which integrates AI with geometric odometry. Utilizing the KITTI dataset's LiDAR data, INAF dynamically adjusts attention weights based on environmental feedback, enhancing the system's adaptability and measurement accuracy. This method advances the precision of both localization and 3D mapping, demonstrating the potential of our fusion technique to enhance autonomous navigation systems in complex scenarios.

\end{abstract}

%%%%%%%%%%%%%%%%%%%%%%%%%%%%%%%%%%%%%%%%%%%%%%%%%%%%%%%%%%%%%%%%%%%%%%%%%%%%%%%%
\section{INTRODUCTION}

Advancements in autonomous navigation systems, particularly for unmanned aerial vehicles (UAVs) and unmanned ground vehicles (UGVs), have been pivotal in addressing the complexities of navigating and mapping environments where Global Navigation Satellite Systems (GNSS) are either denied or unreliable. LiDAR (Light Detection and Ranging) technology has emerged as a cornerstone in this domain, offering high-resolution spatial data crucial for both localization and environmental modeling. However, the effective use of LiDAR data in scenarios devoid of GNSS support requires sophisticated computational techniques that can interpret this data accurately under varying conditions.

The integration of Artificial Intelligence (AI) with geometric algorithms in LiDAR Simultaneous Localization and Mapping (SLAM) presents a promising avenue for enhancing these capabilities. This paper introduces a novel fusion methodology, termed Inferred Attention Fusion (INAF), which leverages the strengths of both AI and traditional geometric approaches to overcome the limitations of existing SLAM methods. By dynamically adjusting the fusion process based on real-time environmental feedback, INAF aims to enhance the accuracy of localization and the richness of 3D maps, facilitating more reliable navigation in GNSS-denied environments.

\begin{figure}[t!]
    \centering
    \includegraphics[width=0.8\columnwidth]{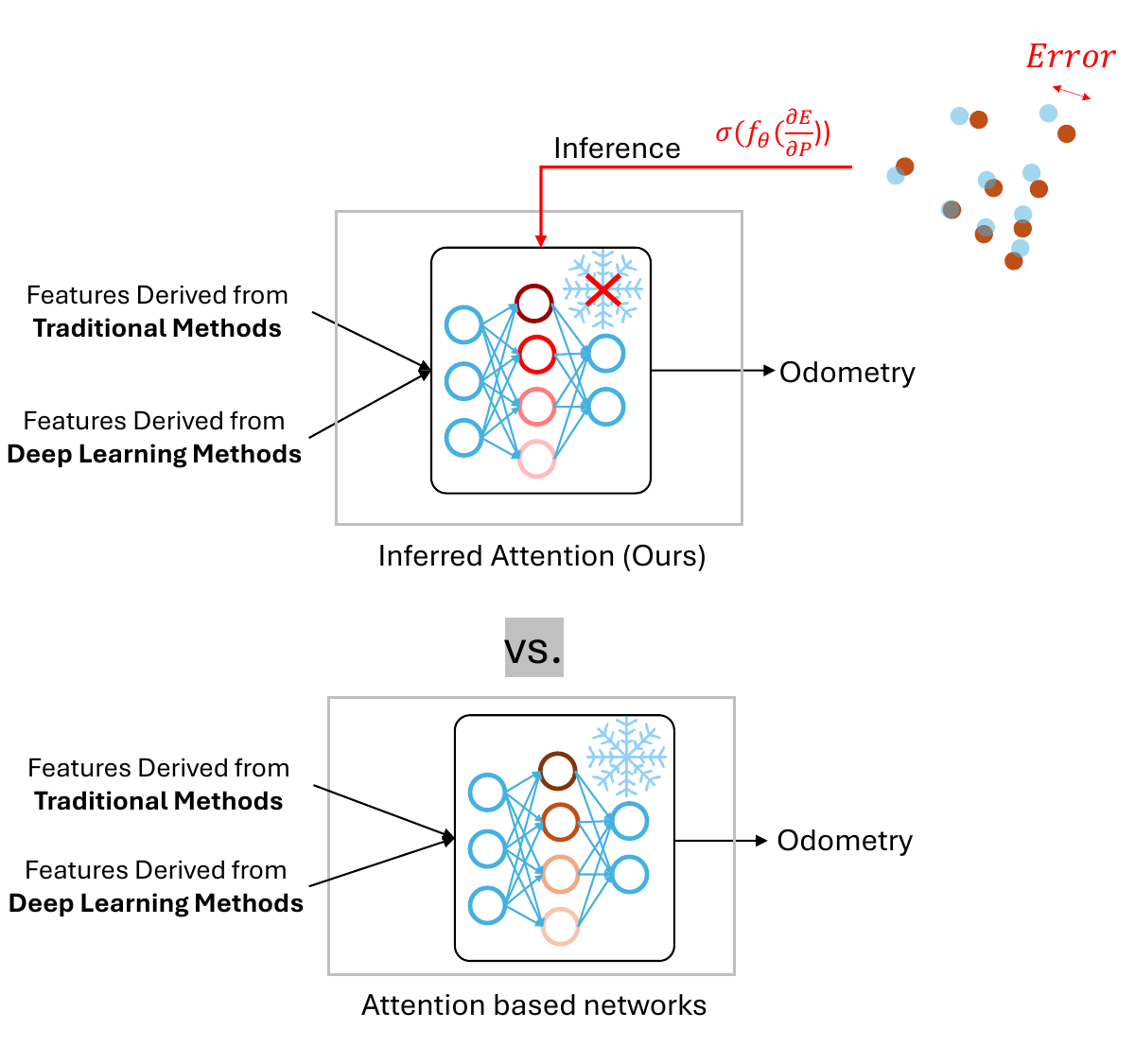}
    \caption{Comparison of network adaptation strategies: General Attention-Based Networks vs. Inferred Attention (Ours). This figure illustrates how each network type adapts to new data. General attention networks use static attention weights that are determined during the training phase and do not change in response to new environmental data. In contrast, our Inferred Attention model dynamically adjusts its attention weights in real-time, utilizing feedback from environmental changes to enhance adaptation and improve accuracy in mapping and localization tasks.}
    \label{fig:fig1}
\end{figure}

\section{RELATED WORK}

This section reviews the progression and integration of methodologies in the domain of Simultaneous Localization and Mapping (SLAM), emphasizing the transition from traditional geometric approaches to the inclusion of deep learning and sensor fusion techniques.

\subsection{Evolution of SLAM Technologies}

SLAM has been a fundamental problem in robotics, aiming to build an accurate map of an environment while simultaneously tracking the robot's location within it. The concept was first introduced by Randall Smith in 1990, who used an Extended Kalman Filter (EKF) for the incremental estimation of a robot's posture and landmark positions \cite{ref_ch6_sotarev}. Durrant-Whyte and Bailey later defined the complete probabilistic formulation for SLAM problems in 2006, coining the term SLAM \cite{ref_ch6_sotarev}.

\subsection{Advances in Sensor Technologies and Architectural Improvements}

With advances in sensor technology, SLAM systems have incorporated a variety of sensors, including LiDAR, cameras, and Inertial Measurement Units (IMUs), enhancing state estimation from traditional filter-based approaches to more sophisticated graph optimization techniques \cite{ref_ch6_sotarev}. Architecturally, the shift from single-threaded to multi-threaded systems has significantly improved the efficiency and scalability of SLAM applications, enabling more robust and dynamic environment mapping and navigation \cite{ref_ch6_sotarev}.

\subsection{Challenges in GNSS-based and Visual SLAM}

Despite the accuracy of GNSS-based SLAM in open environments, its reliability decreases under signal obstruction or reflection, prompting the integration of other sensors to overcome these challenges \cite{ref_ch6_GNNSreview}. Visual SLAM (VSLAM) has become popular due to its low cost and the ease of sensor fusion, although it still faces significant challenges in complex environments \cite{ref_ch6_vslamreview}. Recent efforts in semantic SLAM, which utilizes high-level environmental information, promise to enhance the environmental understanding significantly \cite{ref_ch6_vslamreview}.

\subsection{LiDAR-based SLAM in Complex Environments}

LiDAR-based SLAM is particularly noted for its application in large-scale environments and has been extensively developed across sectors like automated driving and urban surveying \cite{ref_ch6_lidarsota}. However, it can suffer from performance degradation due to dynamic motion or feature-poor settings \cite{ref_ch6_lidarreview}. Recent studies have focused on multi-sensor fusion within LiDAR-based SLAM to address these limitations, showing promising results \cite{ref_ch6_lidarreview}.

\subsection{Integration of Deep Learning in SLAM}

The incorporation of Deep Neural Networks (DNNs) in SLAM, known as DNN-SLAM, represents a significant advancement, particularly in static environment mapping and scene reconstruction \cite{ref_ch6_dns}. These systems, however, still face challenges like tracking drift and mapping errors in dynamic scenarios. Innovative approaches such as feature point segmentation that combines semantic features are being explored to enhance the robustness and accuracy of DNN-based SLAM in real-world applications \cite{ref_ch6_ddnslam}.

\subsection{The Role of Hybrid and Fusion Techniques}

Hybrid SLAM systems that combine classical geometric estimators with learned components are emerging as solutions to the limitations of traditional methods. These systems leverage the strengths of both geometric and machine learning approaches, enhancing the SLAM's adaptability and accuracy in dynamic environments \cite{ref_ch6_loam, ref_ch6_li2019net}.

\section{Methodology}

This section outlines the methodologies adopted for enhancing LiDAR SLAM through the integration of Inferred Attention Fusion (INAF) with traditional geometric and deep learning methods. We detail the design of the odometry pipelines, the rationale behind choosing fusion networks over ensemble learning, and introduce the novel concept of inferred attention.

\subsection{Odometry Pipelines}

\begin{figure}[t!]
    \centering
    \includegraphics[width=1\columnwidth]{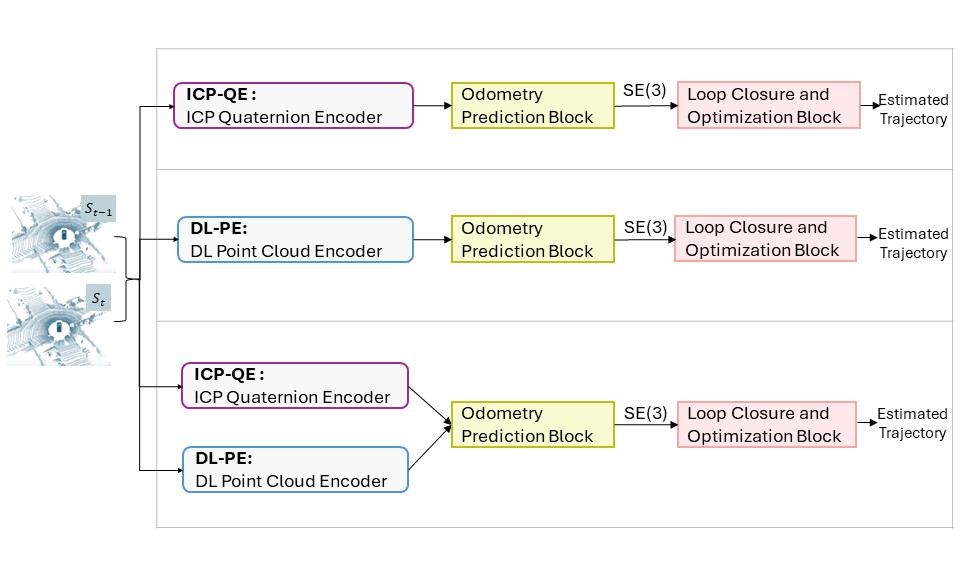}
    \caption{Pipelines illustrating three configurations: two single-branch setups (ICP-QE and DL-PE) and one fused setup combining both ICP-QE and DL-PE branches, followed by the Odometry Prediction Block and the Loop Closure and Optimization Block.}
    \label{ch6_fig:Odom}
\end{figure}

Our study introduces three distinct odometry pipelines to address various aspects of the SLAM process, as illustrated in Figure \ref{ch6_fig:Odom}.

\begin{itemize}
    \item The first pipeline, \textbf{ICP-QE}, utilizes geometric computations between consecutive scans to predict transformations, followed by integration through a loop closure and optimization block.
    \item The second pipeline, \textbf{DL-PE}, employs deep learning techniques to extract and predict motion features from sensor data, again followed by an odometry prediction block and a loop closure and optimization block.
    \item The third pipeline combines the strengths of both ICP-QE and DL-PE methodologies, enhancing the accuracy and robustness of the odometry estimation. This combined approach processes these estimates to refine trajectory estimation for comprehensive mapping and localization tasks.
\end{itemize}

\subsection{Ensemble Learning vs. Fusion Networks}

\begin{figure}[t!]
    \centering
    \includegraphics[width=1\columnwidth]{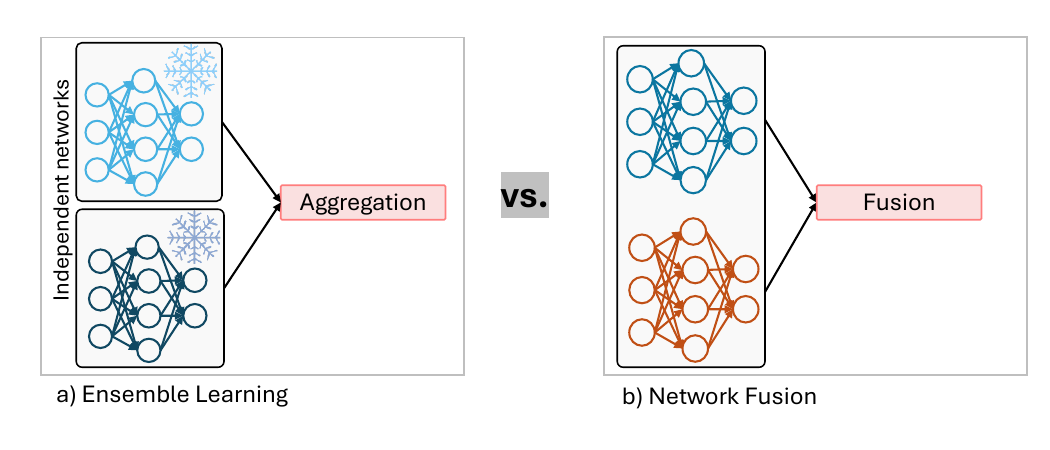}
    \caption{a) Ensemble Learning vs. b) Network Fusion. In ensemble learning, multiple networks are trained independently and remain frozen during combination. In contrast, network fusion involves training multiple networks together, allowing for integrated learning and mutual adaptation.}
    \label{fig:ch6Ens}
\end{figure}

In addressing the integration of multiple data sources, we explore the comparative advantages of ensemble learning and fusion networks, as depicted in Figure \ref{fig:ch6Ens}.

Fusion networks, or sensor fusion networks, are preferred over ensemble learning due to their ability to dynamically integrate and optimize various data sources during training. This method proves especially useful in handling complex and dynamic tasks such as SLAM where continuous adaptation to new data is crucial.

\subsection{Advantages of Inferred Attention Over Incremental and Attention-Based Approaches}

\begin{figure}[t!]
    \centering
    \includegraphics[width=1\columnwidth]{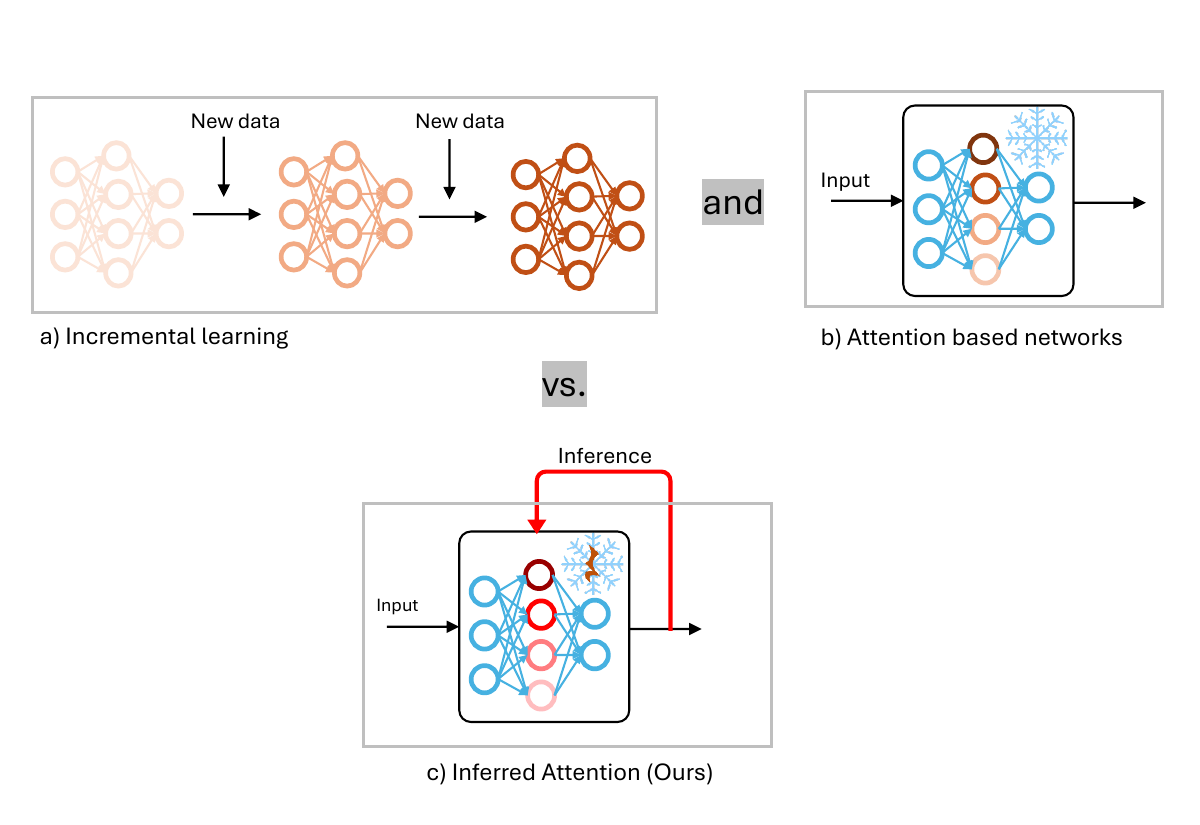}
    \caption{Adaptation of networks to new data across different methods: (a) Incremental learning, where the network is updated with new data over time; (b) Attention-based networks, which utilize attention mechanisms to focus on relevant parts of the input; and (c) Inferred Attention (Ours), where network parameters are continuously adapted based on feedback from the environment.}
    \label{fig:ch6Adapt}
\end{figure}

The novel approach of Inferred Attention is introduced to address the limitations of incremental learning and traditional attention-based methods. Figure \ref{fig:ch6Adapt} showcases the adaptation of network parameters to new data, highlighting the dynamic adjustments based on real-time environmental feedback.

Inferred Attention enhances the robustness and accuracy of the SLAM system by enabling the network to respond effectively to unobserved motion and environmental changes without the need for extensive retraining.

\subsection{Proposed Inferred Attention Fusion}

\begin{figure}[t!]
    \centering
    \includegraphics[width=1\columnwidth]{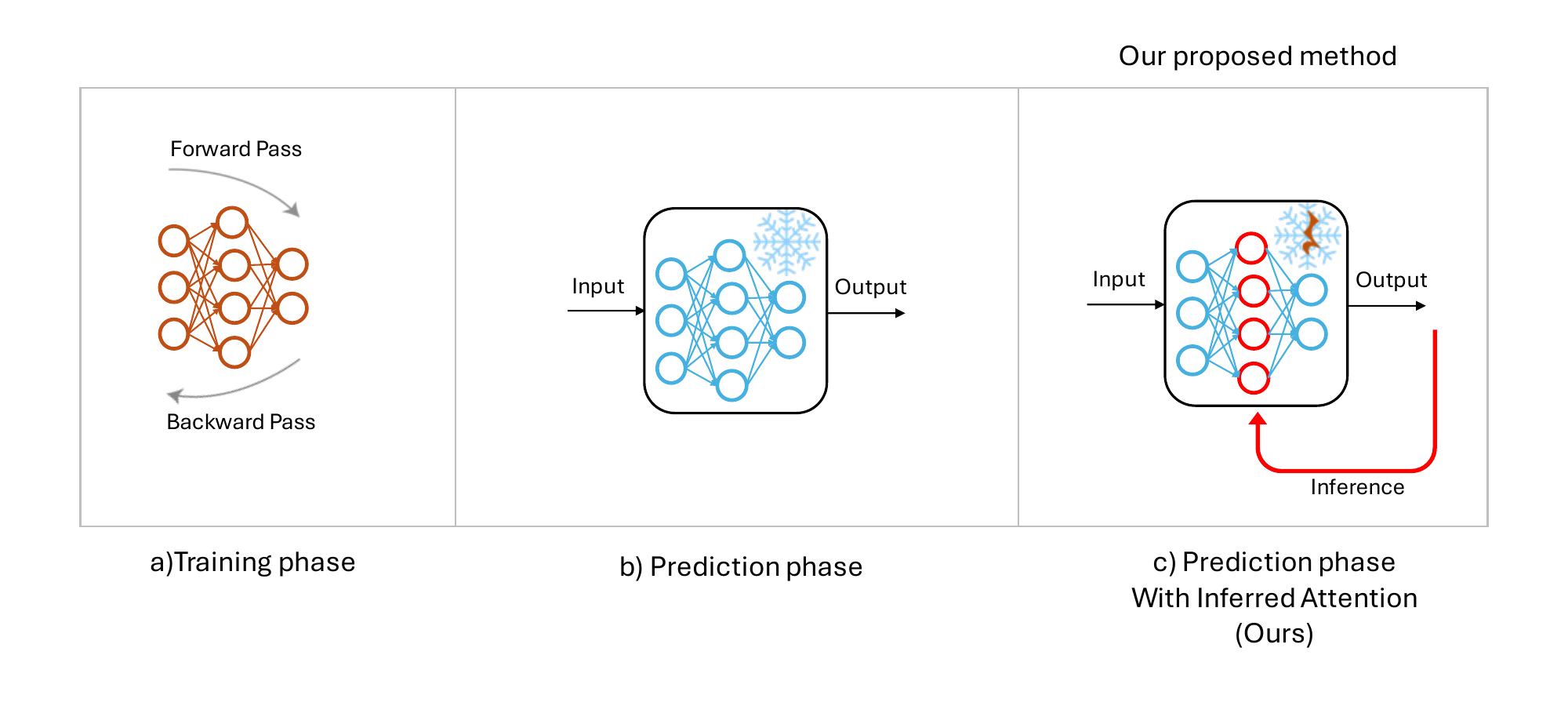}
    \caption{Overview of the three phases in our network under INAF: (a) Training phase, where the network learns from data; (b) Prediction phase, where the trained network is frozen and used for inference, and (c) Prediction phase with Inferred Attention Fusion, where some network parameters continue to adapt based on feedback from the environment.}
    \label{fig:ch6INAF}
\end{figure}

The Inferred Attention Fusion (INAF) approach is designed to allow the network to continue adapting post-training, ensuring high performance even in dynamic and unpredictable environments. Figure \ref{fig:ch6INAF} describes the three phases of the network operation under INAF.

\subsection{Backend Optimization}

Finally, the backend optimization processes involving scan loop closure and graph optimization are crucial for integrating the trajectory estimates into a coherent map and localizing the robot within it. We employ scan context techniques and Minisam for these tasks, ensuring high accuracy and consistency in the SLAM process.

%_________________________________________________New Section
\section{EXPLAINING SYSTEM MODULES}

\begin{figure*}[htbp]
    \centering
    \begin{subfigure}{1.5\columnwidth}
        \centering
        \includegraphics[width=0.85\columnwidth]{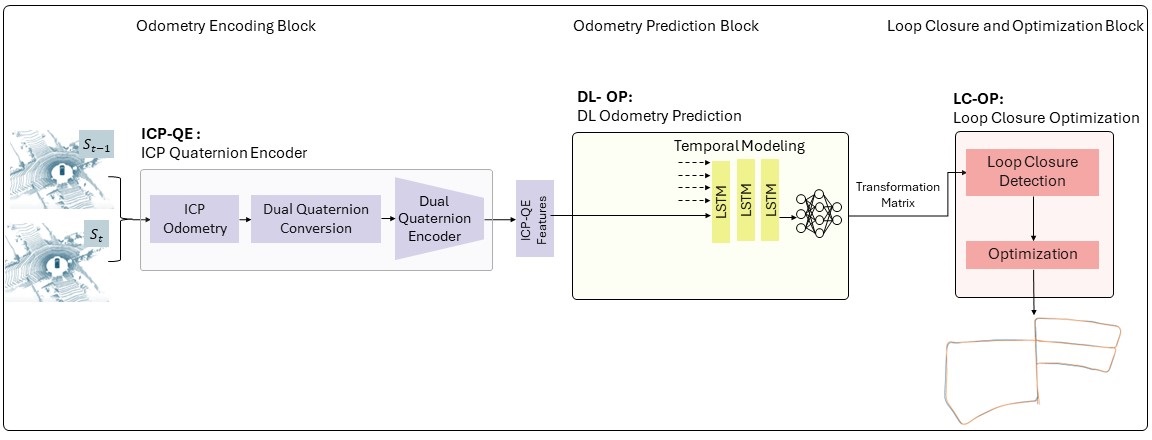} 
        \caption{ICP-QE block followed by the DL-OP module and the loop closure optimization block.}
        \label{fig:first}
    \end{subfigure}
    
    \begin{subfigure}{1.5\columnwidth}
        \centering
        \includegraphics[width=0.85\columnwidth]{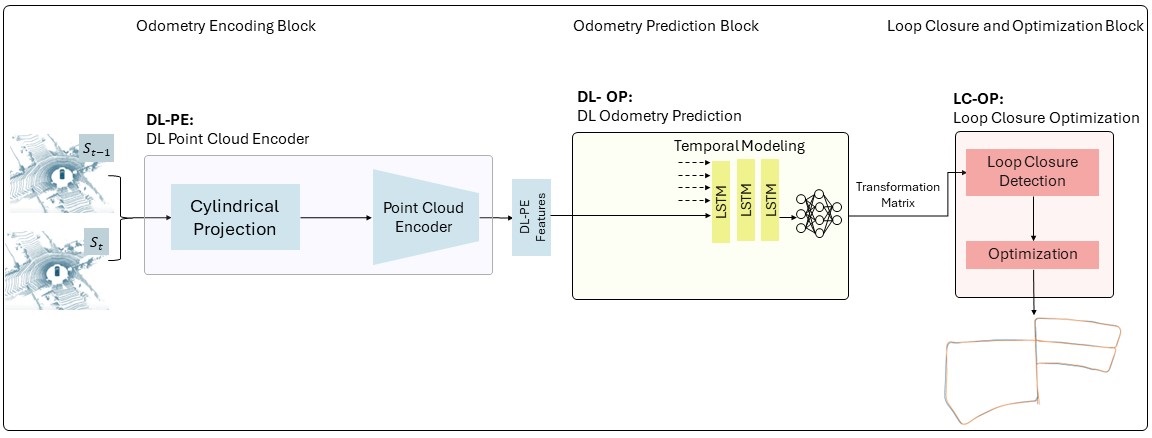} 
        \caption{DL-PE block followed by the DL-OP module and the loop closure optimization block.}
        \label{fig:second}
    \end{subfigure}
    
    \begin{subfigure}{1.5\columnwidth}
        \centering
        \includegraphics[width=0.85\columnwidth]{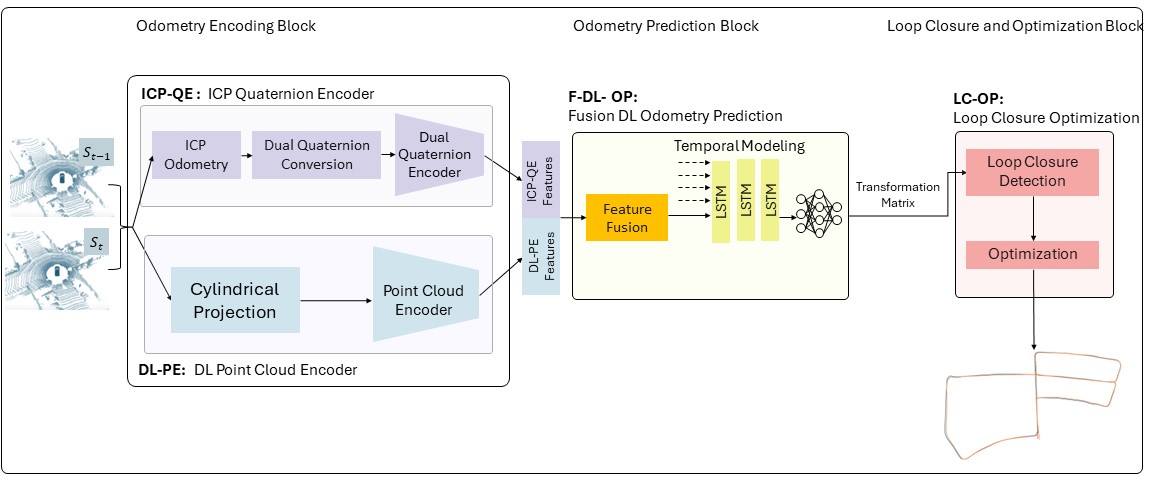} 
        \caption{Fusion of ICP-QE and DL-PE branches, followed by the F-DL-OP module and the loop closure optimization block.}
        \label{fig:third}
    \end{subfigure}
    
    \caption{The proposed SLAM system, illustrating three configurations: (a) Fusion of ICP-QE and DL-PE, (b) ICP-QE and (c) DL-PE followed by the Odometry Prediction and Loop Optimization Blocks}
    \label{ch6_fig6:3net}
\end{figure*}

Our LiDAR SLAM methodology employs three distinct odometry pipelines, each designed around key components: the odometry encoding block, the odometry prediction block, and the loop closure and optimization block. These are depicted in Figure \ref{ch6_fig6:3net} and are designed to leverage the strengths of both geometric and deep learning methodologies to enhance odometry estimation's accuracy and robustness.

\subsection{Odometry Pipelines}

The \textbf{ICP-QE Pipeline} uses a geometric-based approach, starting with the alignment of point clouds through the Iterative Closest Point (ICP) algorithm. The transformations computed are then encoded using dual quaternion encoding to optimize the geometric data representation for subsequent processing. The overall transformation within the Dual Quaternion Encoder is represented as:

\begin{align}
\theta_{\text{ICP-QE}} = f_{\text{Dual Quaternion Encoder}}(& (q_0, q_1, q_2, q_3) \nonumber \\
& + \varepsilon (q_{e0}, q_{e1}, q_{e2}, q_{e3}))
\end{align}

The \textbf{DL-PE Pipeline} leverages deep learning to extract and predict motion features from sensor data, transforming these via a deep convolutional network:
$$
\theta_{DL-PE} = f_{\text{Pointcloud Encoder}}(u, v, p^z, \rho, \iota, N_\xi, N_\upsilon, N_\zeta)
$$

The \textbf{Fused Pipeline} combines outputs from both the geometric and deep learning methods. It integrates these diverse inputs at the odometry prediction stage through a feature fusion module, enhancing overall system accuracy:
$$  
z = g(\theta_{\text{ICP-QE}}, \theta_{\text{DL-PE}})
$$

\subsection{Odometry Prediction Block}

This block, common to all pipelines, includes temporal modeling and task-solving components that use a recurrent neural network (LSTM) to incorporate temporal dependencies essential for accurate odometry estimation:
$$  
y_t = \text{RNN}(z_t, h_{t-1})
$$

\subsection{Loop Closure and Optimization}

This block enhances trajectory estimation by integrating loop closure detections with graph optimization techniques, refining the trajectory and improving map accuracy.

\subsection{Loss Function and Model Optimization}

The model employs a mean squared error (MSE) loss function to optimize the relative pose predictions:
$$  
L(\theta)^2 = \|\hat{p} - p\|^2 + \lambda^2 \|\hat{r} - r\|^2
$$

This function ensures that the SLAM system is capable of handling the real-world operational demands, demonstrating superior performance in a variety of environmental conditions.

\section{FUSION MODULES} \label{ch6_sec_Fusion}
In this section, we explore feature fusion modules, highlighting the limitations of traditional fusion methods that treat all features equally without considering dynamic environmental factors and potential drifts. We introduce the INAF (Inferred Attention Fusion) methodology, a novel dynamically operating module that recalibrates feature importance in real-time based on motion feedback, aimed at adaptively enhancing system accuracy.

\subsection{Fusion Methodologies}
We compare INAF with two traditional fusion methods: Direct Fusion and Attention-Based Fusion, serving as baselines for assessing the improvements offered by INAF.

\textbf{Direct Fusion:} This method integrates features from different sensor modalities through concatenation, employing Multilayer Perceptrons (MLPs) to seamlessly merge information:
\[
g_{\text{direct}}(\theta_{\text{ICP-QE}}, \theta_{\text{DL-PE}}) = [\theta_{\text{ICP-QE}}; \theta_{\text{DL-PE}}]
\]
Here, features \(\theta_{\text{ICP-QE}}\) and \(\theta_{\text{DL-PE}}\) are concatenated into a single feature vector that captures combined modalities.

\textbf{Attention-Based Fusion:} This method uses a multi-head self-attention mechanism to dynamically weight the importance of data points based on their relevance:
\[
\text{Output} = \text{softmax}\left(\frac{(\text{Queries} \cdot \text{Keys}^T)}{\sqrt{d_k}}\right) \cdot \text{Values}
\]
where Queries, Keys, and Values are transformed outputs of the concatenated features, allowing the system to focus on the most pertinent features dynamically.

\subsection{Inferred Attention Fusion (Proposed Method)}
We introduce the Inferred Attention Fusion (INAF) as a dynamically operating module that recalibrates feature importance based on real-time motion feedback. This module employs a series of loss functions to assess and correct for any drifts or errors during feature fusion, significantly enhancing the adaptiveness and accuracy of the fusion process.

The error correction in INAF utilizes three distinct loss functions: point-to-point, point-to-plane, and plane-to-plane. These loss functions enable the module to precisely align transformed source scans with target scans. The equations for these loss functions are given by:
\begin{align}
\mathcal{L}_{\text{point to point}} &= \frac{1}{n_k} \sum_{p=1}^{N} \left| \left( \widehat{\text{transformed source}}_p - \text{target}_p \right) \right|_2^2, \\
\mathcal{L}_{\text{point to plane}} &= \frac{1}{n_k} \sum_{p=1}^{N} \left| (\widehat{\text{transformed source}}_p - \text{target}_p) \cdot \mathbf{n}_p \right|_2^2, \\
\mathcal{L}_{\text{plane to plane}} &= \frac{1}{n_k} \sum_{p=1}^{N} \left| (\mathbf{n} \cdot \widehat{\mathbf{p}} - \mathbf{n}_p) \right|_2^2.
\end{align}

The total error is a summation of these individual errors, calculated as follows:
\begin{equation}
\text{E} = \mathcal{L}_{\text{point to point}} + \mathcal{L}_{\text{point to plane}} + \mathcal{L}_{\text{plane to plane}}
\end{equation}

Using the chain rule, we calculate the gradient of the total error with respect to the branch features \(\theta\):
\begin{equation}
\frac{\partial E}{\partial \theta} = \frac{\partial E}{\partial q} \cdot \frac{\partial q}{\partial \theta}
\end{equation}

Here, \(\theta\) represents the concatenated feature vectors from the ICP-QE and DL-PE branches. A neural network \(f_{\theta}\) is used to map the normalized error rate vector \(e'\) to a set of weights, applying a sigmoid activation function:
\begin{equation}
w = \sigma(f_{\theta}(\frac{e - \mu_e}{\sigma_e}))
\end{equation}
where \(e'\) is the normalized rate of error change, and \(w\) denotes the resulting weights which are applied element-wise to the feature vectors.

These adaptive weights recalibrate the contributions of each feature based on their relative importance in minimizing the fusion error:
\begin{equation}
\theta'_{\text{ICP-QE}} = w \odot \theta_{\text{ICP-QE}}, \quad \theta'_{\text{DL-PE}} = w \odot \theta_{\text{DL-PE}}
\end{equation}
where \(\odot\) denotes element-wise multiplication.

\begin{figure}[t!]
\centering
\includegraphics[width=1\linewidth]{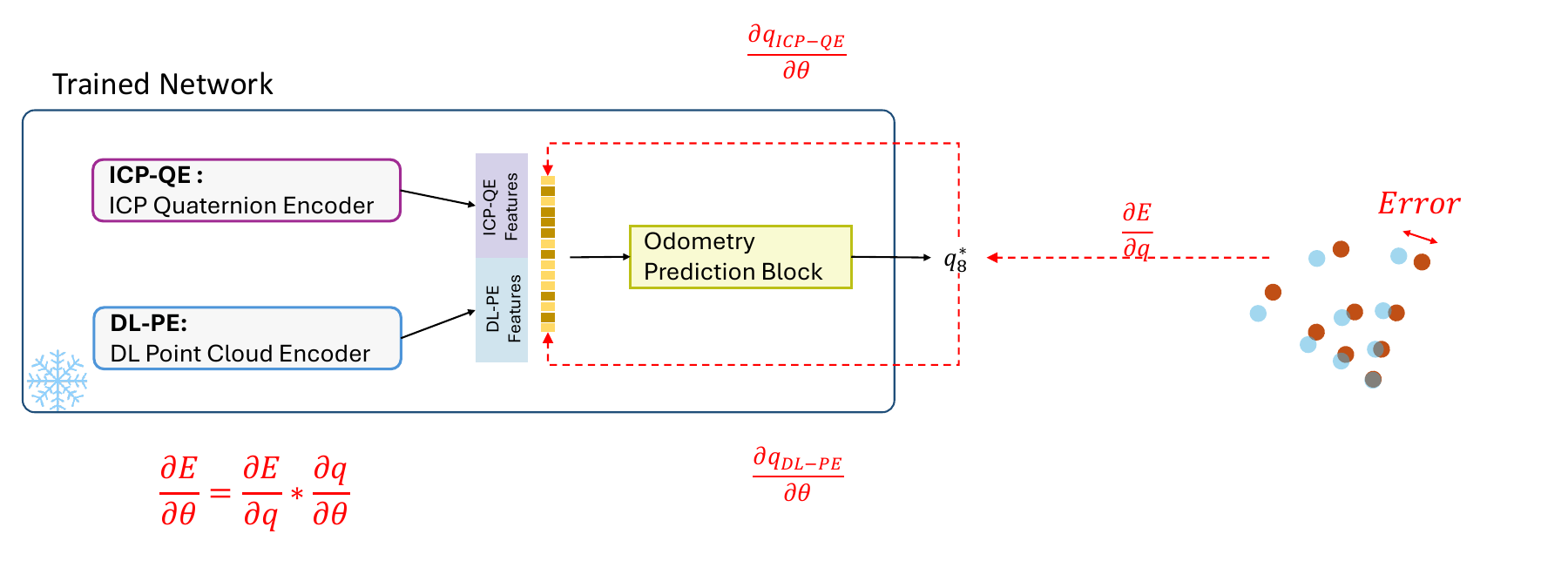}
\caption{Illustration of the chain rule in calculating error gradients for the INAF fusion method, tracing back from the output error to the features derived from the branches \(\theta_{\text{ICP-QE}}\) and \(\theta_{\text{DL-PE}}\).}
\label{fig:ch6math1}
\end{figure}

This innovative approach ensures that the fusion module continuously adapts to changing environmental conditions, significantly improving the robustness and accuracy of the system in dynamic scenarios.

\section{Training Procedure}
The training of our network is meticulously structured to optimize performance and generalization. The procedure involves data preparation, optimization strategy implementation, and rigorous model validation.

\subsection{Data Preparation and Splitting}
Our dataset is split into training and validation subsets with a 80/20 ratio, respectively. Data shuffling ensures a random distribution of samples, which helps in maintaining the integrity of the model's ability to generalize. This setup is critical in mitigating the risk of overfitting and effectively validating the model's performance on unseen data.

\subsection{Optimization Strategy}
We employ the Adam optimizer due to its efficiency in dealing with sparse gradients and its adaptability to large datasets. An exponential decay schedule for the learning rate starts at 0.001 and reduces by a factor of 0.9 every 10,000 steps, which helps in refining the learning process to avoid minima overshoots and enhance convergence.

\begin{figure}
    \centering
    \includegraphics[width=0.7\columnwidth]{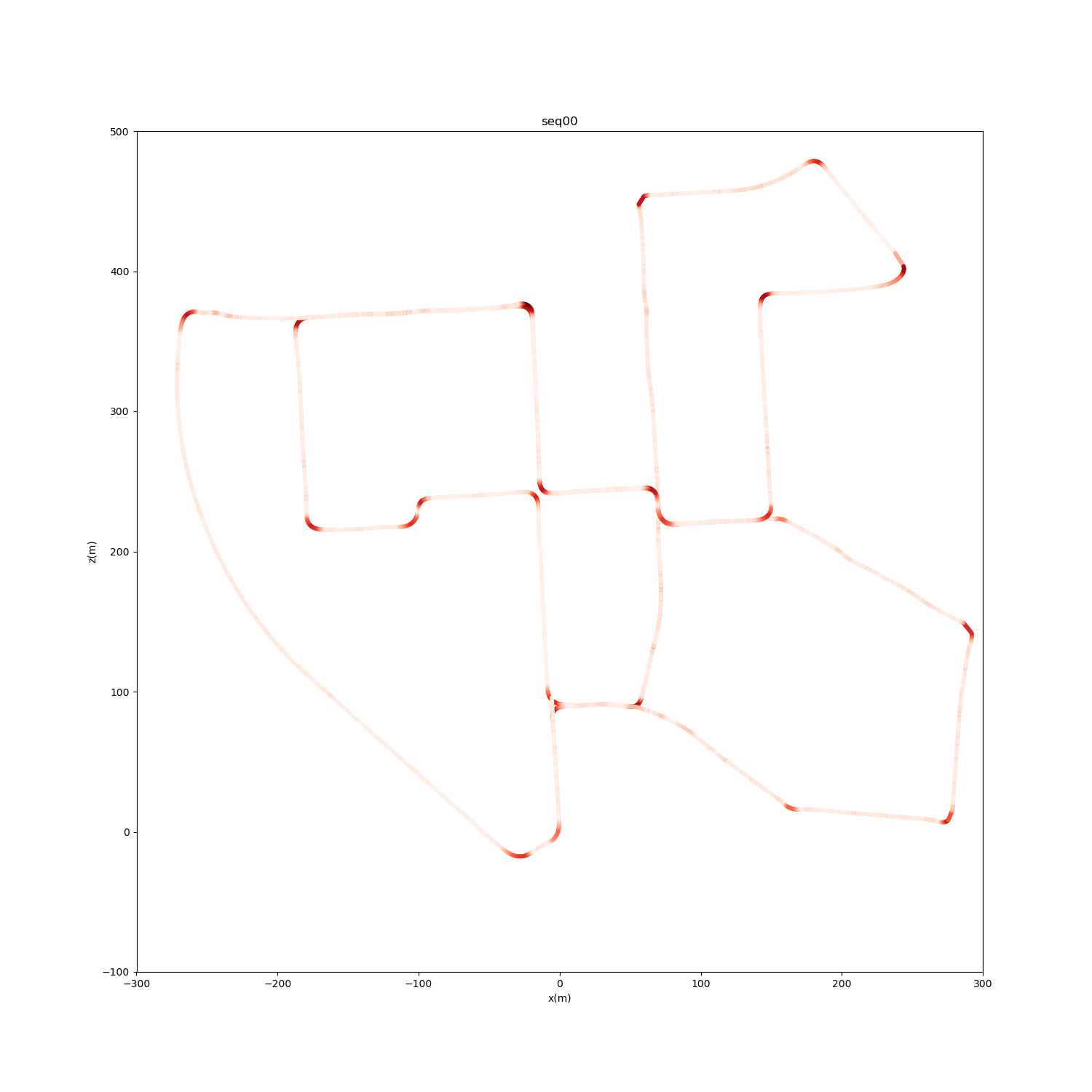}
    \caption{Sequence 00 highlighting large angle rotations.}
    \label{fig:ch6_rot_red}
\end{figure}

\subsection{Custom Loss Function}
Our model utilizes a combined loss function to balance the precision and sensitivity to outliers effectively, which is expressed as:

\begin{align}
L_\text{combined\_loss} = & (1 - \alpha) \times \text{MAE}(y_{\text{true}}, y_{\text{pred}}) \nonumber \\
& + \alpha \times \text{MSE}(y_{\text{true}}, y_{\text{pred}})
\end{align}

Here, \( \alpha \) is set to 0.8, optimizing the trade-offs between Mean Absolute Error (MAE) and Mean Squared Error (MSE).

\subsection{Early Stopping and Model Checkpointing}
Early stopping halts training after 15 epochs without improvement in validation performance, optimizing computational resources and curbing overfitting. Model checkpointing is implemented to save the state of the best-performing model iteration, ensuring that we retain the most efficient version of the model for further evaluation and operational deployment.

\subsection{Hyperparameter Tuning}
Hyperparameter tuning is approached through the Hyperband algorithm, which efficiently manages computational resources and hastens the optimal configuration search. It dynamically allocates resources to promising configurations and swiftly eliminates less promising ones. The tuning process explores parameters such as batch size, learning rate, and the architecture's depth and activation functions, crucial for the model's performance on unseen data. Our implementation uses the Keras Tuner library to navigate through the hyperparameter space, aiming to minimize validation mean absolute error and ensure the model's generalization:

\begin{equation} \label{eq:heperparam1}
n = \left\lceil \frac{s_{\max} + 1}{s + 1} \cdot \eta^s \right\rceil 
\end{equation}

\begin{equation} \label{eq:heperparam2}
r = \frac{R}{\eta^s}
\end{equation}

\section{Backend Loop Closure and Optimization}
Our system integrates advanced loop closure and graph optimization techniques to enhance performance and accuracy. Loop closure identifies revisited locations and adjusts for accumulated errors, while graph optimization refines the trajectory estimations to minimize errors, leveraging frameworks like miniSAM for efficient large-scale optimization.

\subsection{Loop Closure}
Using 3D LiDAR scans, the scan context technique serves as a novel global descriptor that is robust against LiDAR viewpoint changes, making it invaluable for consistent loop detection and mapping optimization.

\subsection{Graph Optimization}
The miniSAM framework, a part of our optimization toolkit, uses a factor graph approach to model and solve nonlinear least squares problems effectively, optimizing the system's trajectory estimates with high computational efficiency and accuracy.

%________________________

\section{Results}
\subsection{Performance Evaluation}
The relative pose error is a critical metric for assessing trajectory precision over a specified time interval \(\Delta\), essential for evaluating visual odometry. It is computed at time step \(i\) as:

\begin{equation} \label{eq:ch6_Ei}
    E_i = (Q_i^{-1} Q_{i+\Delta})^{-1} (P_i^{-1} P_{i+\Delta})
\end{equation}
and the Root Mean Square Error (RMSE) across different intervals $\Delta$:
\begin{equation} \label{eq:ch6_RMSE1}
    \text{RMSE}(E_{1:n}, \Delta) = \left( \sum_{i=1}^{m} \frac{1}{m} ||\text{trans}(E_i)||^2 \right)^{1/2}
\end{equation}
\begin{equation} \label{eq:ch6_RMSE2}
    \text{RMSE}(E_{1:n}) = \frac{1}{n} \sum_{\Delta=1}^{n} \text{RMSE}(E_{1:n}, \Delta)
\end{equation}

\subsection{Comparison with Classical SLAM}
% We compare the proposed neural network-based method (ICP-QE) with the traditional ICP method, demonstrating substantial improvements in both rotational and translational accuracy. For detailed results, see Table \ref{table:icp_comparison}.

This section examines various odometry estimation techniques, including the traditional Iterative Closest Point (ICP) method and a neural network-based approach utilizing the ICP-Quality Estimation (ICP-QE) branch (Figure \ref{ch6_fig6:3net}.a).

Traditional methods like ICP, known for aligning successive LiDAR scans to estimate vehicle motion, generally show lower orientation errors compared to learning-based methods, which excel in translation error rates by directly extracting features from raw point clouds. Although ICP is straightforward, relying on basic geometric principles to align 3D shapes, it struggles in environments with sparse point clouds or insufficient geometric features.

The neural network method enhances the traditional ICP by incorporating its outputs for initial alignment and using deep learning to refine these estimates. This hybrid approach, particularly the integration of the ICP-QE branch, leverages ICP's initial alignment capabilities while the neural network corrects residual errors, achieving better accuracy in both rotational and translational metrics. The ICP algorithm itself, established in the early 1990s and refined for efficient 3D shape registration in various applications like computer vision and robotics, serves as a fundamental comparison point to demonstrate the potential improvements offered by integrating AI techniques. For detailed results, see Table \ref{table:icp_comparison}.

\subsection{Comparison with AI SLAM}
% The DL-PE module, which bypasses explicit ICP alignment, demonstrates superior rotational motion estimation but requires ICP for precise translational estimations. Refer to Table \ref{table:icp_comparison} for a comparative analysis.
The second deep learning (DL)-based method, the DL-PE module, processes point cloud data by initially using Cylindrical Projection before inputting it into the network. This method eliminates the need for explicit Iterative Closest Point (ICP) alignment, relying solely on the neural network's ability to discern and leverage features from the point cloud data. In terms of rotational motion estimation, the LiDAR branch outperforms both the traditional ICP method and the geometric branch. This enhancement stems from the network's adeptness at capturing complex spatial relationships within the point cloud, which is essential for accurate rotational estimates. However, for translational error, the LiDAR branch does not surpass the geometric branch that incorporates ICP input. This indicates that while the neural network effectively processes point cloud data for rotational dynamics, the initial alignment from ICP remains crucial for precise translational motion estimation. For a comparative analysis, refer to Table \ref{table:icp_comparison}.

\begin{table}
\caption{Comparative performance of different odometry estimation methods.}
\renewcommand{\arraystretch}{1.1} % Adjust vertical padding
\setlength{\tabcolsep}{4pt} % Adjust horizontal padding
\resizebox{\columnwidth}{!}{%
\begin{tabular}{||c|c|c||cc||cc|}
\hline
\multirow{2}{*}{Methods} & 
\multirow{2}{*}{\begin{tabular}[c]{@{}c@{}}Feature \\ Weighting\end{tabular}} & 
\multirow{2}{*}{\begin{tabular}[c]{@{}l@{}}Back-\\ End\end{tabular}} & 
\multicolumn{2}{c|}{Training Set} & 
\multicolumn{2}{c|}{Test Set}\\ 
\cline{4-7} &  &  & 
\multicolumn{1}{c|}{\begin{tabular}[c]{@{}c@{}}Trans. Error\\ (\%)\end{tabular}} & 
\begin{tabular}[c]{@{}c@{}}Rot. Error\\ (deg/100m)\end{tabular} & \multicolumn{1}{c|}{\begin{tabular}[c]{@{}c@{}}Trans. Error\\ (\%)\end{tabular}} & 
\begin{tabular}[c]{@{}c@{}}Rot. Error\\ (deg/100m)\end{tabular} \\ \hline
ICP & -- & -- & -- & -- & 101.23 & 31.32 \\ \hline
ICP-QE & Direct & Yes & 5.37 & 4.22 & 6.31 & 6.83 \\ \hline
DL-PE & Direct & Yes & 5.21 & 4.07 & 5.85 & 3.86 \\ \hline
\end{tabular}%
}
\label{table:icp_comparison}
\end{table}

\begin{table}
\caption{Performance comparison of various methods on KITTI dataset.}
\centering
\setlength{\tabcolsep}{4pt} % Adjust horizontal padding
\resizebox{\columnwidth}{!}{
\begin{tabular}{|l|c|c|c|c|}
\hline
Method & \multicolumn{2}{c|}{Training Sets 00-08} & \multicolumn{2}{c|}{Test Sets 09 and 10} \\ \cline{2-5} &
\multicolumn{1}{c|}{\begin{tabular}[c]{@{}c@{}}Trans. Error\\ (\%)\end{tabular}}  & 
\begin{tabular}[c]{@{}c@{}}Rot. Error\\ (deg/100m)\end{tabular} &
\multicolumn{1}{c|}{\begin{tabular}[c]{@{}c@{}}Trans. Error\\ (\%)\end{tabular}} &
\begin{tabular}[c]{@{}c@{}}Rot. Error\\ (deg/100m)\end{tabular} \\ \hline
\multicolumn{1}{|c|}{\begin{tabular}[c]{@{}c@{}}INAF Fusion +\\ Loop Closure\end{tabular}} & 2.15 & 0.82 & 2.26 & 0.92 \\ \hline
Velas et al. & 2.94 & NA & 4.11 & NA \\ \hline
DeLORA & 3.00 & 1.38 & 6.25 & 2.58 \\ \hline
ICP & NA & NA & 101.23 & 31.32 \\ \hline
LOAM & NA & NA & 1.36 & 0.51 \\ \hline
\end{tabular}
}
\label{tab:sota_comparison}
\end{table}

\subsection{Advanced Fusion Techniques}
% We explore the efficacy of INAF Fusion, comparing it with direct and attention-based fusion methods. The INAF Fusion significantly enhances performance, especially in rotational accuracy. For performance under different conditions, see Table \ref{tab:sota_comparison}.

The INAF fusion method outperformed direct and attention-based fusion methods in translation and rotation error metrics, both with and without loop closure optimization. Loop closure, a common technique for correcting drift in odometry and SLAM systems, significantly improved accuracy in all methods by reducing accumulated errors over time.

Without loop closure, the direct fusion method showed a translation error of 3.92\% and a rotation error of 2.61° per 100 meters, which dropped to 2.81\% and 1.82° with loop closure. The attention-based method performed better, achieving 3.42\% and 2.13° without loop closure, and improving to 2.73\% and 1.41° with it.

The INAF method delivered the best performance, with translation and rotation errors of 2.93\% and 1.18° in the training set without loop closure, reduced to 2.15\% and 0.82° with loop closure. Test set errors improved from 5.72\% and 2.01° without loop closure to 2.26\% and 0.92° with it. These results demonstrate INAF's advanced fusion capabilities, which dynamically adapt to varying environmental conditions, providing superior accuracy and stability in real-world scenarios. See Table \ref{tab:sota_comparison} for performance under different conditions.

\subsection{Adaptive Sensor Fusion}
% The INAF fusion system dynamically adjusts to diverse driving scenarios, effectively managing large-angle rotations and rapidly changing environments. Figure \ref{fig:ch6_rot_red} illustrates these capabilities.

\section{Adaptive Sensor Fusion in Straight-Driving Estimation Using INAF}

Our investigation into straight-driving translation estimation reveals a dynamic interplay between sensor modalities, reflecting the adaptive nature of the INAF fusion system. The weighting of modalities such as ICP and LiDAR shifts depending on driving conditions, underscoring their complementary strengths. Specifically, the ICP branch often dominates in scenarios that emphasize precise driving dynamics, while the LiDAR branch plays a critical role in accurately capturing straight-line motion under varying conditions.

In environments featuring rapid rotations or moving objects, the LiDAR-based branch demonstrates heightened influence. This suggests that AI-extracted features are particularly effective in addressing the challenges posed by dynamic environments. This adaptive weighting mechanism highlights the INAF system's ability to optimize performance by aligning modality importance with scenario-specific dynamics.

\begin{figure}[htbp]
  \centering
  \resizebox{0.7\columnwidth}{!}{ % Scale the whole figure to fit the column width
    \begin{tabular}{c} % Stack subfigures vertically using a single column
      \begin{subfigure}{\textwidth}
        \centering
        \includegraphics[width=\textwidth]{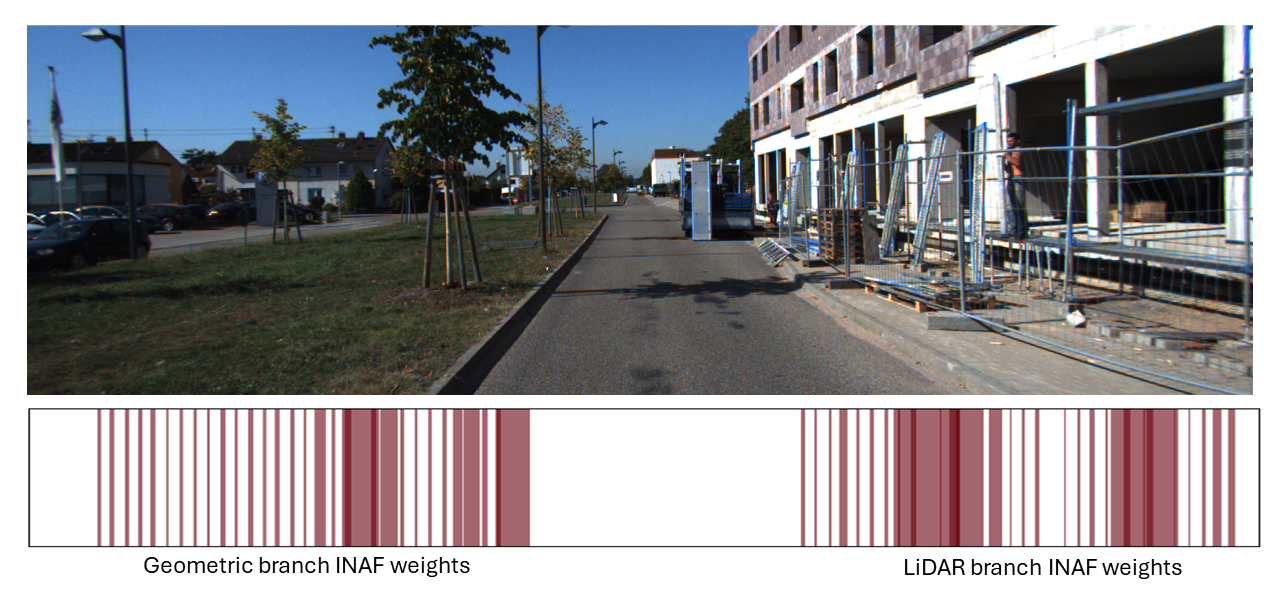}
        \caption{Driving straight}
        \label{fig:ch6_mask:sub1}
      \end{subfigure} \\
      
      \begin{subfigure}{\textwidth}
        \centering
        \includegraphics[width=\textwidth]{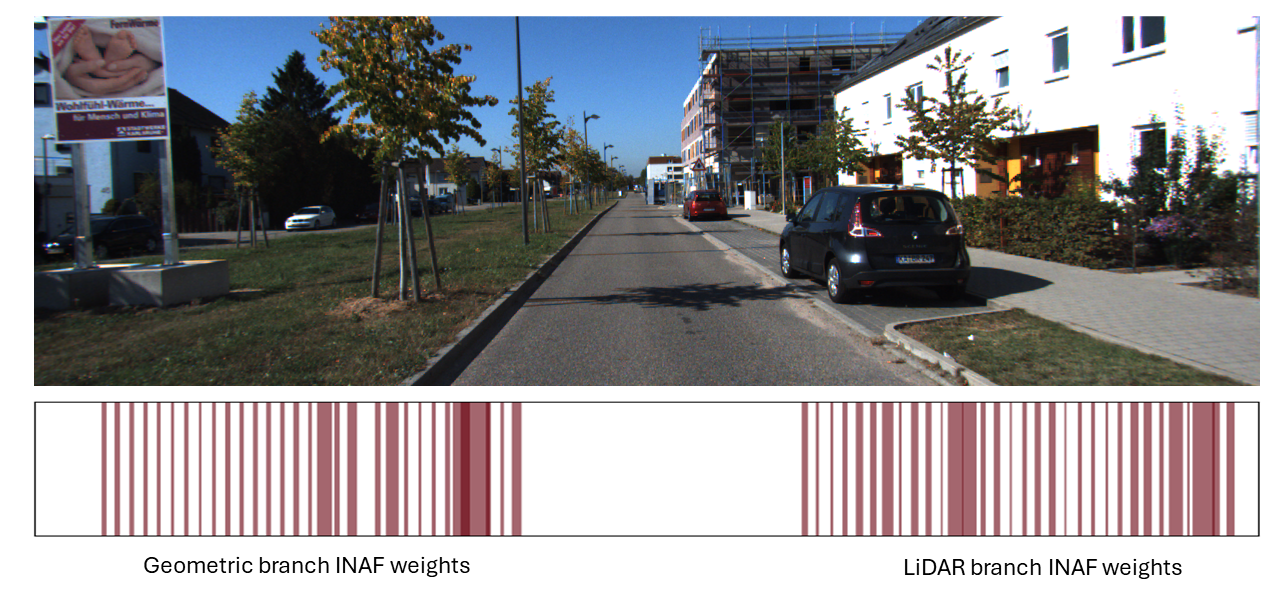}
        \caption{Driving straight}
        \label{fig:ch6_mask:sub2}
      \end{subfigure} \\
      
      \begin{subfigure}{\textwidth}
        \centering
        \includegraphics[width=\textwidth]{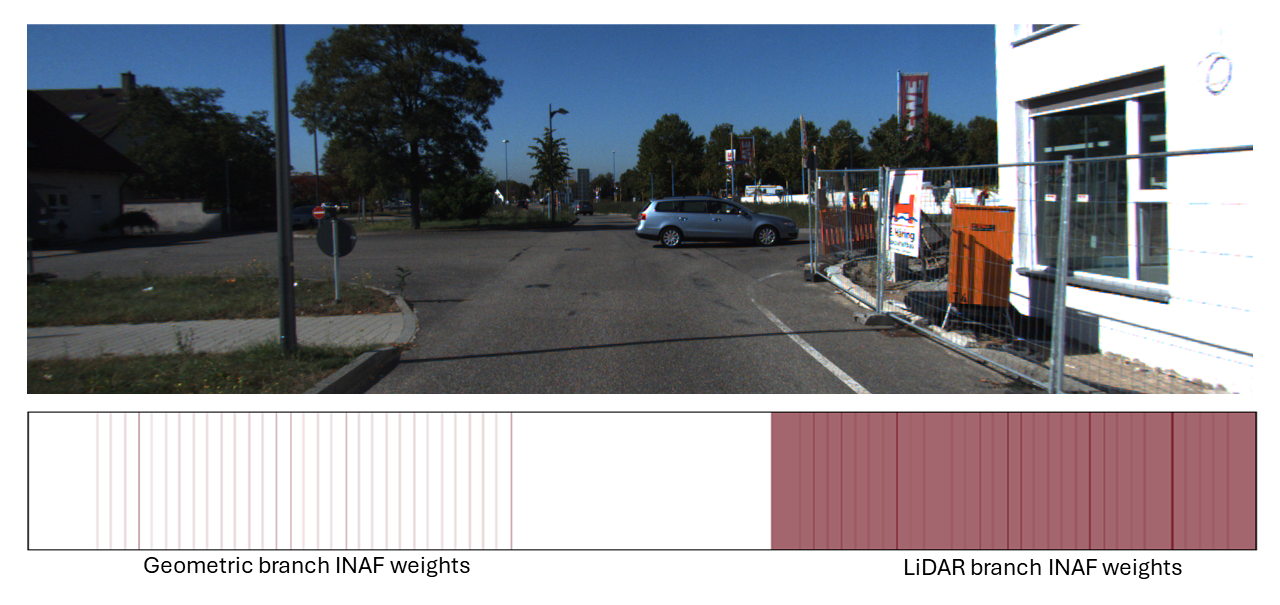}
        \caption{Moving object}
        \label{fig:ch6_mask:sub3}
      \end{subfigure} \\
      
      \begin{subfigure}{\textwidth}
        \centering
        \includegraphics[width=\textwidth]{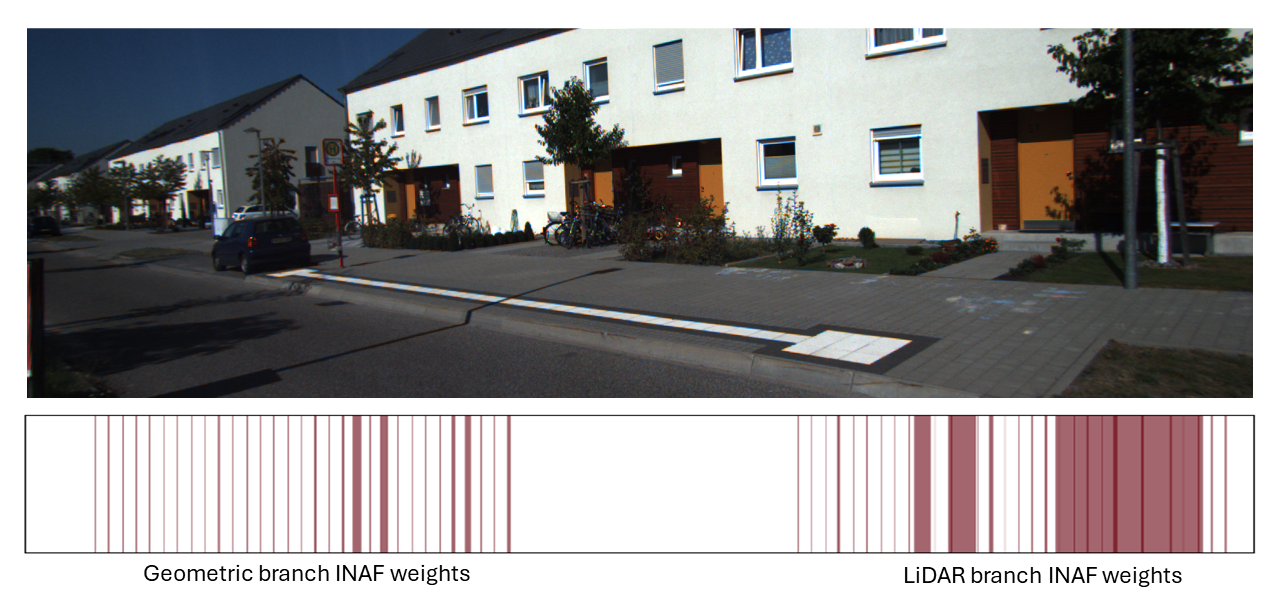}
        \caption{Turning}
        \label{fig:ch6_mask:sub4}
      \end{subfigure}
    \end{tabular}
  }
  \caption{Visualization of learned weights for INAF fusion across different scenarios: (a) straight driving, (b) the presence of a moving object (a car), and (d) a turning maneuver. The weights of selected features from the geometric and LiDAR branches demonstrate the adaptive nature of the fusion process, with varying importance assigned to each branch depending on the scenario.}
  \label{fig:ch6_mask}
\end{figure}

Comprehensive experiments conducted on the KITTI dataset validate these findings across diverse driving scenarios. By employing visualization techniques to analyze the sensor fusion masks, as illustrated in Figure ~\ref{fig:ch6_mask}, we reveal a strong correlation between the selected features of each branch and the dynamics of the environment or measurements, further supporting the robustness of the proposed approach.

This analysis not only deepens our understanding of the unique contributions of each modality but also provides actionable insights for designing future multimodal systems. These systems will benefit from enhanced adaptability and performance in real-world driving conditions.

Figure \ref{fig:ch6_mask} visualizes the learned weights under different conditions, demonstrating the INAF fusion's adaptability.

\section{SUMMARY OF CONTRIBUTIONS}
The primary contributions of this work are:
\begin{itemize}
    \item The introduction of Inferred Attention Fusion (INAF), a dynamic fusion strategy for real-time feature weighting in LiDAR SLAM.
    \item The integration of geometric (ICP-QE) and deep learning (DL-PE) odometry pipelines within a unified architecture.
    \item A comprehensive evaluation of INAF against classical and AI-based SLAM methods, demonstrating superior accuracy and adaptability.
\end{itemize}

\section{CONCLUSION}

This paper introduced a novel LiDAR-based SLAM framework that integrates geometric and deep learning odometry pipelines through the proposed Inferred Attention Fusion (INAF) module. By dynamically recalibrating feature importance based on real-time environmental feedback, INAF enhances the adaptability, robustness, and precision of localization and mapping in GNSS-denied environments.

Experimental results on the KITTI dataset demonstrated that INAF outperforms conventional direct and attention-based fusion techniques, significantly reducing both translational and rotational errors. The method also showed strong resilience to dynamic motion and feature-sparse scenarios, indicating its effectiveness for complex real-world applications. 

Furthermore, the adaptive weighting mechanism of INAF reveals that the system intelligently balances the contributions of geometric and AI-driven components depending on motion dynamics and environmental context. This adaptability provides a powerful pathway for generalizing SLAM performance across varied operational conditions.

Future work will focus on extending INAF to multi-sensor SLAM systems that include visual and inertial modalities, further improving robustness under extreme motion or perception noise. Integration with online learning and cross-domain adaptation strategies will also be explored to enable fully autonomous recalibration during long-term deployment.

Overall, INAF establishes a promising foundation for next-generation autonomous navigation systems that can learn, adapt, and optimize in real time.

%________

% \section{CONCLUSIONS}

% A conclusion section is not required. Although a conclusion may review the main points of the paper, do not replicate the abstract as the conclusion. A conclusion might elaborate on the importance of the work or suggest applications and extensions. 

\addtolength{\textheight}{-6cm}   % This command serves to balance the column lengths

%%%%%%%%%%%%%%%%%%%%%%%%%%%%%%%%%%%%%%%%%%%%%%%%%%%%%%%%%%%%%%%%%%%%%%%%%%%%%%%%

%%%%%%%%%%%%%%%%%%%%%%%%%%%%%%%%%%%%%%%%%%%%%%%%%%%%%%%%%%%%%%%%%%%%%%%%%%%%%%%%



% \section*{ACKNOWLEDGMENT}

% The preferred spelling of the word ÒacknowledgmentÓ in America is without an ÒeÓ after the ÒgÓ. Avoid the stilted expression, ÒOne of us (R. B. G.) thanks . . .Ó  Instead, try ÒR. B. G. thanksÓ. Put sponsor acknowledgments in the unnumbered footnote on the first page.

%%%%%%%%%%%%%%%%%%%%%%%%%%%%%%%%%%%%%%%%%%%%%%%%%%%%%%%%%%%%%%%%%%%%%%%%%%%%%%%%

\end{document}